# Cobb Angle Measurement of Scoliosis with Reduced Variability

Raka Kundu, Amlan Chakrabarti
A. K. Choudhury School of Information Technology,
University of Calcutta,
Kolkata, India.
kundu.raka@gmail.com, acakcs@caluniv.ac.in

Prasanna K. Lenka
Department of Rehab Engineering,
National Institute for the Orthopaedically Handicapped
Kolkata, India.
lenka_pk@yahoo.co.uk

*Abstract*— Cobb angle, which is a measure of spinal curvature, is the standard method for quantifying the magnitude of Scoliosis, related to spinal deformity in orthopedics. Determining the Cobb angle through manual process is subject to human errors. In this work, we propose a methodology to measure the magnitude of Cobb angle, which appreciably reduces the variability related to its measurement, compared to the related works. The proposed methodology is facilitated by using a suitable new improved version of Non-Local Means (*NLM*) for image denoisation and Otsu's automatic threshold selection [14, 15] for Canny edge detection. We have selected NLM for preprocessing of the image as it is one of the fine states of art for image denoisation and helps in retaining the image quality. Trimmed-mean, median are more robust to outliners than mean and following this concept we observed that NLM denoising quality performance can be enhanced by using Euclidean trimmed-mean replacing the mean. To prove the better performance of the Non-Local Euclidean Trimmed-mean (*NLETM*) denoising filter, we have provided some comparative study results of the proposed denoising technique with traditional NLM and Non-Local Euclidean Medians (*NLEM*) [11]. The experimental results for Cobb angle measurement over intra-observer and inter-observer experimental data reveals the better performance and superiority of the proposed approach compared to the related works. MATLAB2009b image processing toolbox was used for the purpose of simulation and verification of the proposed methodology.

*Keywords-Scoliosis; Cobb angle, Computer aided measurement, Non-Local Euclidean Trimmed-mean, Otsu's threshold.*

I. INTRODUCTION

Scoliosis [1] is a three dimensional deformity of spine that involves abnormal lateral curve accompanied with vertebral rotation. Scoliosis is categorized into several types depending on the cause and age of the curve development. 2% to 3% of the population can be affected by scoliosis [2]. The most common form of Scoliosis is idiopathic scoliosis where the cause is unknown. Based on the curve condition, treatment of scoliosis involves observation, bracing and surgery. The golden standard method in orthopedics for evaluating degree of lateral curvature of scoliosis is measurement of Cobb's angle from anterior–posterior spine X-ray image. Cobb angle measurement helps in realizing the stage of deformity, supervise chances of curve progression and management of scoliosis. Manual Cobb angle measurement process from X-ray plate involves selection of the extreme vertebrae, which incline more severely towards the concavity of the spinal curvature and drawing lines from the upper endplate of the superior vertebrae and the lower endplate of the inferior vertebrae of the curve. The angle *(θ)* formed between these two lines is the Cobb angle (Fig. 1).The general cause of measurement errors in Cobb angle are for selection of different end vertebrae and due to estimation of improper slope of the end vertebrae. Even same selected end vertebra may result in Cobb angle degree variation. This happens when there is inaccurate drawing of lines across the endplates of the selected vertebra. It requires nearly about 20 minutes on an average for the manual measurement process.

In such a situation computerized Cobb angle measurement becomes an effective tool for Scoliosis evaluation and reducing the variability of Cobb angle measurement. Several computerized Cobb angle measurement techniques have been developed to improve the reliability and accuracy of measurement method till date. In reference [3] the authors have developed a computer aided Cobb angle quantification method that produced eight lines on the region of interest (ROI) and resulted in eight equal segments. There was a need of marking the two points on each line, where they intersect the vertebra edge. The process computed midpoint of every line and outlined the vertebral column midline connecting these midpoints. Cobb angle was calculated based on this midline. The process is tedious as it needs much manual intervention. The authors in reference [4] have proposed an automatic Cobb angle measurement method based upon active shape model, but the method required training set. A reliability assessment of Cobb angle using manual and digital method was carried out in reference [5]. The digital method included land marking on the superior and inferior endplates of curve's extreme vertebrae. Marks were needed on the two extremities of each endplate and rest of the steps for Cobb angle quantification was done automatically. Here improper placement of the landmarks on the extreme corners of the endplates increases the chances of Cobb angle degree variation. A new technique of digital Cobb angle measurement was proposed in reference [6], where contrast and brightness adjustment of the selected ROI was performed. Canny edge detection, fuzzy Hough transform was used to identify endplate lines of the vertebrae. Magnitude of the Cobb angle was computed based on the slope of these lines. However this Cobb angle measurement method has more measurement variability than the proposed method of this paper.

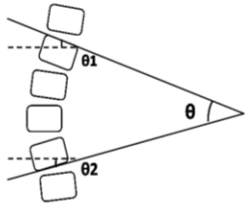

Figure 1: Cobb angle (θ = θ1+θ2)

This paper proposes a computer-assisted method for Cobb angle measurement using image processing techniques. Our specific contributions are as follows:
- Implementation of Euclidean trimmed-mean for NLM denoising filter which acts relatively insensitive to outliners and works expeditiously for noise removal of digital X-ray images. The performance of the filter in terms of preserving the image quality is appreciable.
- Suitable application of the traditional Otsu's thresholding method for automatic and proper threshold selection from the image. This help in improving the performance of Canny edge detection by retraining the fake edges.
- Overall reduction in variability of Cobb angle measurement compared to the best known related work [5, 6].

The proposed technique of Cobb angle measurement aims in reducing user intervention by proper identification of vertebral end-plate slope, which minimizes the measurement variability of Cobb angle.

Remainder of this paper is organized as follows. Section II describes the improved denoising technique and the image processing algorithms used in our proposed scheme of Cobb angle measurement. Experimental data and comparative experiments of the proposed method are illuminated in section III. Conclusion and scope of our future work are discussed in Section IV.

## II. PROPOSED METHODOLOGY

The vertebral column is a complex structure and in many cases the vertebra shape in an X-ray image is not consistent. So it becomes difficult to process the whole image. In the proposed methodology the extreme vertebrae, which tilt more towards the concativity of the curve, is initially selected by the user. The Cobb angle is the sum of the angles of the extreme-vertebrae (Fig. 1). After selecting the region of interest we have used NLETM denoising technique which facilitates the post processing of the image. Next, histogram equalization was performed for contrast enhancement. The process continued with Canny edge detection where Otsu method was used for automatic threshold selection. The choice of threshold value is based on the histogram of the image, which makes the concept of threshold selection more logical. For detecting the slope of the extreme vertebrae edge detection was followed by Hough transform. Cobb angle was calculated from the detected slopes of the extreme vertebrae. Fig. 2 explains the flow diagram of our proposed scheme.

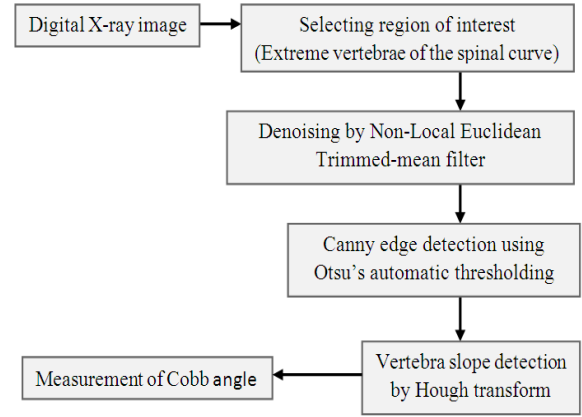

Figure 2: Flow diagram of Cobb angle measurement technique.

### A. Denoising by Non-Local Euclidean Trimmed-mean

*1) Non-Local Means Filter(NLM):* The Non-Local Means filtering [7] technique reflects an admirable concept of image denoisation. It is developed on self similarity of texture. Considering three pixels $p$, $q_1$, $q_2$ of a given image (Fig. 3), the neighborhood of pixels $p$ and $q_1$ are more similar than the neighborhood of the pixels $p$ and $q_2$. This self similarity concept can be used for denoising an image. So, $q_1$ will have more weight for the denoising value of pixel $p$ in comparison to $q_2$.

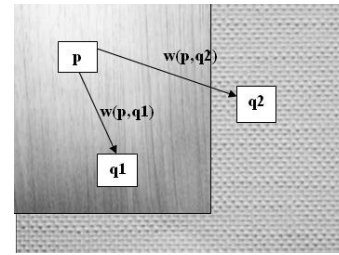

Figure 3: Non local means strategy

*2) Procesing steps in NLM:* The weighted average function for NLM is as follows:

$$NLMv(p) = \sum w(p,q)\, v(p) \qquad (1)$$

Here, each restored pixel value $v(p)$ in the image is the weighted average of all the pixels in the image. $w(p,q)$ is the weight and $v(q)$ represent other pixels of the image. The weights $w(p,q)$ are based on the similarity between the neighborhood of pixels $p$ and $q$.

$$w(p,q) = \frac{1}{z(p)} e^{\frac{-(d(p,q_n))}{h^2}} \qquad (2)$$

$1/z(p)$ is the normalizing constant, $h$ is the weight-decay control parameter. When $h$ is small there will be less blurring and vice versa and $d(p,q)$ is the weighted squared euclidean difference between neighborhood of pixel $p$ and neighborhood of pixel $q$.

For evaluating a normalizing constant we use the following equation:

$$Z(p) = \sum_{n=1}^{r*s} e^{\frac{-(d(p,q_n))}{h^2}} \quad (3)$$

Also the equation for weighted squared euclidean difference is:

$$d(p,q) = (v(p) - v(q))^2 \quad (4)$$

Where, $(v(p)-v(q))^2$ is the squared euclidean difference between neighborhood of pixel $p$ and neighborhood of pixel $q$. Thus the NLM formula to restore the denoised value of pixel $p$ from the image is:

$$NLMv(p) = \sum w(p,q)\, v(p)$$
$$= \sum \frac{1}{z(p)} e^{\frac{-(d(p,q_n))}{h^2}} v(q) \quad (5)$$

If $N$ denotes the number of pixels in each dimension of the image, $R$ x $R$ denote the neighborhood patch size with $R=(2i+1)$ centered around pixel $p$ and $q_n$. The traditional NLM method has a complexity of $O(N^4.R^2)$. As the chances of similar neighborhood of pixel $p$ and $q$ generally lies in the local neighborhood [8] around the current position of $p$, the traditional NLM algorithm can be simplified by restricting the search space of pixel $q$ to a local reighborhood region of size $SxS$, where $S=2s+1$, around the pixel $p$ instead of the whole image. Hence, the complexity of the algorithm will be $O(N^2.S^2.R^2)$ where $S$ is very very less than $N$.

*3) Non-Local Euclidean Trimmed-mean:* Trimmed-mean is calculated by discarding certain percentage of the lowest and highest scored data and computing the mean from the remaining scores. The equation is as follows

$$Tm = \frac{1}{n-2k} \sum_{i=k+1}^{n-k} x_i \quad (6)$$

Where, $x_i = \{x_1, x_2, x_3, ..... x_n\}$ is the observed scores and $k$ is the trimming percentage. From statistics [9,10] it is know that trimmed-mean is relatively insensitive to outliners than mean. Using this concept we have implemented trimmed-mean for NLM replacing the mean.

*4) Better robustness using Trimmed-mean:* To prove the better performace of NLETM we have experimented with two pixels from two different X-ray images, one at noise level sigma=10 and other at noise level sigma=20. The values of Table 1 prove the efficiency of our proposed technique for estimating true pixel intensity from noisy X-ray images.

TABLE I. ESTIMATION OF INTENSITY BY DENOISING FILTERS

| Noise | Original | Noisy | NLM | NLEM | NLETM |
|---|---|---|---|---|---|
| $\sigma=10$ | 0.52 | 0.48 | 0.49 | 0.50 | 0.53 |
| $\sigma=20$ | 0.53 | 0.45 | 0.49 | 0.51 | 0.53 |

*5) Contrast Enhancement and Edge Detection by Otsu's Thresholding:*

After the process of noise removal, Histogram equalization is performed over the X-ray image for contrast enhancement [13]. It enhances the global contrast of images, especially when the pixel intensity of the image is represented by close contrast values. The range of pixel intensity is 0 to 255 in gray scale images. Through Histogram equalization, the pixels can be better distributed on the whole intensity range of the histogram. This allows image pixels of lower local contrast to gain higher contrast and can be useful in images that have both foregrounds and backgrounds, either bright or both dark or in a very close intensity range. It results in better view of the bone structure in X-ray images and facilitates proper threshold selection for edge detection.

Edge detection is a prerequisite before Hough transform, it gives important information about the basic shape of the image. Canny edge detection is performed based on Otsu's automated threshold selection method [14, 15], which increases the chances of proper edge detection effect.

*B. Hough Transform*

Hough transform [16, 17] is a method for detecting straight lines. $x_i cos\theta + y_i sin\theta = \rho;$ is general representation of a line in x-y plane where, $\rho$ is the distance between the line and the origin, $\theta$ represents the vector angle from the origin to the nearest point on the line. $\rho$-$\theta$ plane is considered as Hough space and $(x_i, y_i)$ is a point on the line that crosses through a certain point $(\rho_k, \theta_k)$ in $\rho$-$\theta$ plane. Points $(x_1, y_1)$, $(x_2, y_2)$ in the x-y plane that passes through a common point $(\rho^{'}, \theta^{'})$ of the Hough space always belong to same line. If $n$ numbers of points lie on the same line $L$ in x-y plane, then there are $n$ sinusoidal curves (every curve represent as a particular point in x-y plane) in Hough space, which passes through a common point $(\rho^{'}, \theta^{'})$. Therefore identification of a line that passes through a bundle of points in x-y plane is decreased to identification of a point of intersection of the sinusoidal curves in the Hough space.

Now, end plates of the vertebra represent the longest line in the vertebra edge detected image. Selecting the line with maximum group of $(x_i, y_i)$ points that passes through a particular $(\rho_k, \theta_k)$ will help in calculating the slope of the vertebra from the edge detected image.

*C. Measurement of Cobb Angle*

Hough transform calculates the slopes from respective edge detected image of upper extreme vertebra and lower extreme vertebra and finally the required Cobb angle $(\theta = \theta_1 + \theta_2)$ is calculated as the sum of the obtained angles from the two extreme vertebra of the spinal curve.

## III. RESULTS AND DISCUSSION

### A. Experimental Study on Denoisation.

To access the performance of the NLETM we have introduced stimulated noise in X-ray image. The standard metric Peak signal to noise ratio *(PSNR)* is used for evaluating the image denoisation technique. When pixels are represented as 8 bit (gray scale image), signal represented as 255, the PSNR can be defined as the ratio between the maximum power of the signal and corruption that affects the signal *(image)* representation. PSNR is the measure of peak error. A greater value of PSNR will indicate a good quality image processing.

$$PSNR = 10 \, log \frac{255^2}{MSE} \qquad (7)$$

The performance of the proposed NLETM is compared with the traditional NLM and NLEM at different noise levels *sigma(σ)=10* to *100*. The weight decay parameter *h*, is taken as *10σ* with *patch size=3* and the local search neighborhood is considered as *SxS=21x21*. The selection of this standard settings is proposed in [12]. For NLETM we have trimmed the values by *30%*. The comparative results of PSNR are illustrated in Fig. 4 and Table 2 respectively.

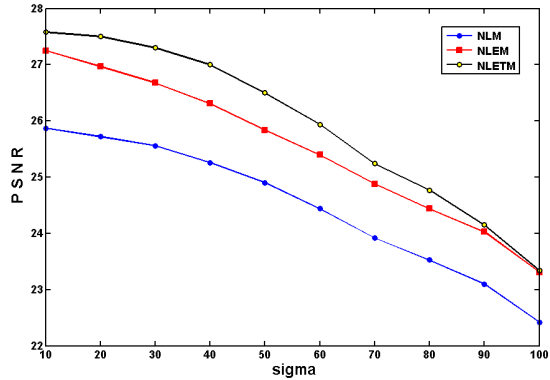

Figure 4. Graphical plot of PSNR showing comparative study of NLM, NLEM, NLETM filtering techniques over radiograph images at noise level sigma (σ)=10 to 100.

TABLE II. PEAK SIGNAL TO NOISE RATIO

| Filters | σ = 10 | σ = 20 | σ = 30 | σ = 40 | σ = 50 |
|---|---|---|---|---|---|
| NLM | 25.87 | 25.72 | 25.56 | 25.26 | 24.9 |
| NLEM | 27.25 | 26.97 | 26.68 | 26.31 | 25.84 |
| NLETM | 27.58 | 27.5 | 27.30 | 27 | 26.5 |
| Filters | σ = 10 | σ = 20 | σ = 30 | σ = 40 | σ = 50 |
| NLM | 24.44 | 23.92 | 23.53 | 23.1 | 22.42 |
| NLEM | 25.4 | 24.88 | 24.44 | 24.03 | 23.31 |
| NLETM | 25.94 | 25.24 | 24.77 | 24.15 | 23.34 |

From the graphical plot and tabulated results of PSNR we see that, the denoising quality of NLETM filtering technique is superior to NLM and NLEM. This suitably justifies our claim for using NLETM as the desired filter for denoising. Next, a number of original noisy X-ray images and its denoised outputs are shown in Fig. 5 for visual verification.

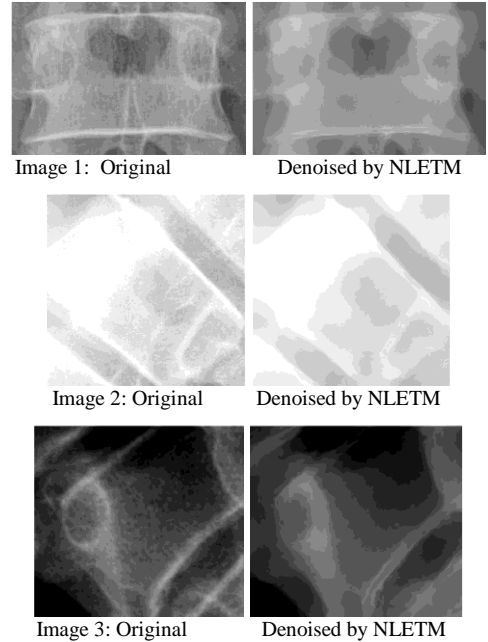

Image 1: Original    Denoised by NLETM

Image 2: Original    Denoised by NLETM

Image 3: Original    Denoised by NLETM

Figure 5. Original X-ray image and its correponding denoised image. Denoisation is performed by NLETM filtering technique.

The out come of this study assures that NLETM filter is efficient enough for cleaning noise over both continuous and discontinous regions of the image. It actively preservers the fine structures, the edges of the X-ray image and gives a quality result.

### B. Results on Edge Detection

Canny edge detection is performend over original noisy X-ray image and over denoised *(NLETM)* X-ray image. To test the filter's impact on improvement of edge detection a visual comparative study (Fig. 6) is carried out where we observe that the edge detection performed after denoisation is more effective. The true edges of the image are identified and the filter helps in proper feature extration from the images. So this makes the use of NLETM filter prior to Canny edge detection meaningful.

Also a comparative study is pursued for experimental affirmation of Otsu's threshold selection on Canny edge detection. In Fig. 7 we have presented some results of Canny edge detected image without and with Otsu threshold application. The Canny edge detected image processed with Otsu's threshold proves relatively suitable for detecting the meaningful edges of vertebra and it will further help in proper vertebral slope detection.

| a | b |
|---|---|
| c | d |

**a:** Original noisy image.
b: Edge detection without denoisation
c: Denoisation with NLETM
d: Edge detection on denoised image.

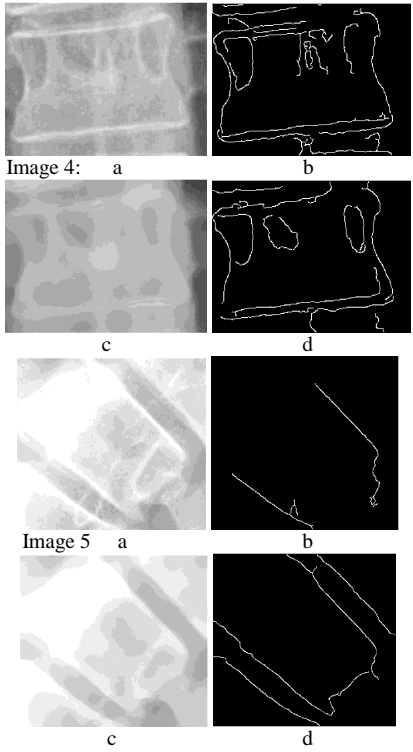

Figure 6. Comparative study showing improvement of edge detection after using fast NLETM denoising technique.

| a | b |
|---|---|

**a:** Canny edge detection without Otsu
**b:** Canny edge detection with Otsu

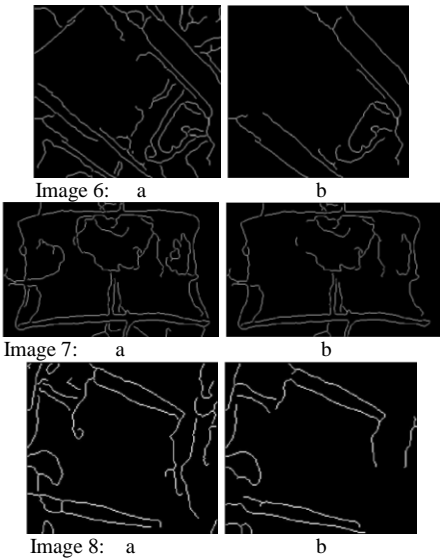

Figure 7. Canny edge detection without Otsu and with Otsu.

## C. Results of Vertebral Slope Detection and Cobb Angle Measurement

Fig. 8 displays an image sequence where vertebral slope is detected automatically by Hough transform for Cobb angle quantification. Table 3 and Table 4 show the reported statistical data of intra-observer measurement variability and inter-observer measurement variability for manual and digital method. In the assessment test of manual and proposed digital Cobb angle measurement technique, the spine angulations was categorized into four groups G1 ($<10^0$), G2 ($10^0$-$25^0$), G3 ($25^0$-$40^0$), G4 ($>40^0$) respectively. and the examination was accomplished by two examiners A1 and A2.

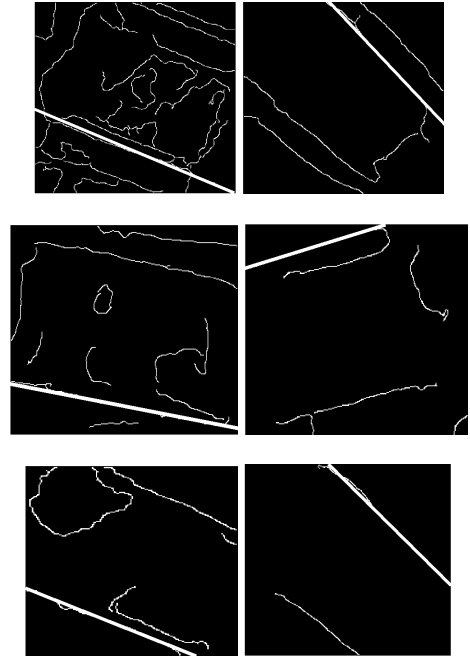

Figure 8. Vertebral slope detected image

TABLE III. INTRA OBSERVER VARIABILITY

| Group | No. | A1 Manual MAD | A1 Digital MAD | A2 Manual MAD | A2 Digital MAD |
|---|---|---|---|---|---|
| G1 | 10 | $1.84^0$ | $1.6^0$ | $1.8^0$ | $2.2^0$ |
| G2 | 7 | $1.9^0$ | $0.9^0$ | $1.4^0$ | $0.1^0$ |
| G3 | 7 | $2.9^0$ | $2.7^0$ | $2.2^0$ | $1.4^0$ |
| G4 | 6 | $2.3^0$ | $1.7^0$ | $2.1^0$ | $1.1^0$ |

TABLE IV. INTER OBSERVER VARIABILITY

| Group | No. | Manual MAD | Digital MAD |
|---|---|---|---|
| G1 | 10 | $1^0$ | $0.6^0$ |
| G2 | 7 | $1.4^0$ | $1.2^0$ |
| G3 | 7 | $1.7^0$ | $1.1^0$ |
| G4 | 6 | $1.5^0$ | $1.1^0$ |

TABLE V. COBB ANGLE ANALYSIS BY DIFFERENT TECHNIQUES

| Metric | Proposed | [6] | [5] |
|---|---|---|---|
| MAD(Intra) | $1.5^0$ | $2.72^0$ | $2.79^0$ |
| MAD(Inter) | $1^0$ | $3.62^0$ | $3.61^0$ |

In Table 3 and Table 4 the analysis of measurement variation is expressed as mean absolute deviation *(MAD)* over series *(No.)* of digital X-ray images. The MAD values as shown in Table 3 and 4 are calculated based on the given formula:

$$MAD = \frac{\sum |x - X|}{n} \quad (9)$$

where, *X*= mean value ; *n* = number of values

The tabulated data of Table 3, evaluates $1.5^0$ and $2.06^0$ as mean intra-observer deviation angle for digital method and manual method considering all scoliosis curvature image groups. Similarly the mean inter-observer deviation angle, evaluated from the data in Table 4, for both digital and manual method is computed as $1^0$ and $1.4^0$ respectively, considering all scoliosis curvature image groups. The comparisons demonstrate that the proposed digital method works better for minimizing measurement variability compared to the manual method. Table 5 gives the comparison of Cobb angle variability measure of the proposed method with the other existing methods [5, 6]. The results in Table 5 show that the MAD value of intra-observer measurement has 44.85 % less variability as compared to the best existing technique [6]. Similarly for the inter-observer measurement, we obtain 72.29 %. less variability compared to the best exiting technique [5]. Hence the results of the proposed method resounds a good impact to discard a step of user intervention by automatic slope selection from vertebra image and also the process produces a very small Cobb angle degree measurement variability.

## IV. CONCLUSION

In this work we have proposed a methodology for Cobb angle measurement, which shows reduced variability compared to the existing research works. Several experiments have been performed on test digital X-ray images and the results demonstrate the effectiveness of our technique. A new Euclidean Trimmed-mean concept for NLM filter is introduced that reduces the affect of outliner and enhances the preprocessing step. In future, we want to improve this technique and extend the scoliosis parameter extraction process by developing the 3D analysis of Scoliosis.